\documentclass[11pt]{article}

\usepackage{amsmath, amssymb}
\usepackage{graphicx}
\usepackage{booktabs}
\usepackage{multirow}
\usepackage{siunitx}
\usepackage{float}
\usepackage{url}
\usepackage{caption}
\usepackage{natbib}
\usepackage{comment}
\usepackage{float}

\usepackage{authblk}
\usepackage[top=1.8cm,bottom=2.2cm,left=2.5cm,right=2.5cm]{geometry}
\title{Precipitation Downscaling Using\\Foundation Model-Conditioned Diffusion}

\author[1]{Victor Nascimento Ribeiro}
\author[1]{Jorge Guevara}
\author[2]{Jorge Sebastián Moraga}
\author[2]{Chris Lucas}
\author[2]{Natalie Lord}
\author[3]{Andrew Taylor}
\author[3]{Edward Lockhart}
\author[1]{Will Trojak}
\author[1]{Johannes Schmude}
\author[1]{Anne Jones\thanks{Corresponding author: anne.jones@ibm.com}}

\affil[1]{IBM Research}
\affil[2]{Fathom}
\affil[3]{STFC Hartree Centre}

\begin{document}

\maketitle
\begin{abstract}
High-resolution precipitation fields are essential for hydrological impact assessment, yet global climate model outputs are too coarse and biased to be used directly. AI-based statistical downscaling with diffusion models offers a promising approach, but the mechanism by which large-scale atmospheric predictors condition the generative process remains largely unexplored. We investigate three conditioning strategies for a denoising diffusion probabilistic model applied to daily precipitation downscaling: channel concatenation of upsampled coarse predictors, cross-attention conditioning with a learned convolutional encoder, and cross-attention conditioning with the frozen encoder of the Prithvi WxC weather foundation model pretrained on global reanalysis. All strategies are evaluated against an unconditioned baseline under identical training and inference conditions, using a comprehensive set of probabilistic, distributional, spectral, and extreme-event metrics for the Colorado River Basin.

Concatenation conditioning achieves the lowest pointwise CRPS and MSE, but tends to produce oversmoothed fields that suppress high-intensity events. In contrast, cross-attention conditioning provides substantially better precipitation distribution realism and modest improvements in spectral fidelity. Improvements are greatest for extremes: the Prithvi-WxC-conditioned model retains over half of $>100$mm/day events, although estimates are uncertain due to limited samples. When trained on the full dataset, the learned convolutional model performs similarly to the foundation model-conditioned approach while requiring lower computational resources. However, the Prithvi-WxC-conditioned model achieves comparable performance with only five years of training data, demonstrating improved data efficiency. These results indicate that cross-attention conditioning offers advantages over simple concatenation for probabilistic precipitation downscaling, and that pretrained foundation model representations may offer further advantages in data-limited settings.

\end{abstract}

\section{Introduction}

Future projections of precipitation time series are needed for risk assessment of climate change impacts such as flooding through simulation with hydrological models. Due to the coarse resolution and biases in the limited representation of small scale effects and extremes, precipitation outputs of GCMs are not suitable for flood modeling. Instead, downscaling methods are employed to derive high-resolution precipitation projections from low-resolution GCM inputs. Statistical downscaling, which is based on observed relationships between large scale climate variables and local climate, has long been established as a computationally efficient and effective downscaling approach, particularly when large ensembles are needed for impact modeling \citep{maraun2010precipitation}. 

Machine learning (ML) has a long history of being applied to this task \citep{Vandal2019, Gutierrez2019}. More recently, a number of authors have explored Deep Learning (DL) applied to downscaling climate projections. Although these methods often offer performance improvements over previous approaches, they are typically deterministic and tend to produce over-smoothed predictions, limiting their ability to represent extremes, uncertainty, and fine-scale spatial variability \citep{Adewoyin2021}. \citet{Rampal2024} provide a review and highlight diffusion as a promising new technique for generating high quality downscaling predictions. 

Diffusion models in AI emerged as an alternative to earlier image generation models such as Generative Adversarial Networks (GANs), which were successful but often difficult to train and prone to instability and "mode collapse" \citep{WGAN}. Diffusion has been used in previous downscaling studies with promising results. \citet{Mardani2025} used residual corrective diffusion to downscale ERA5 reanalysis to 2km radar-assimilated regional weather simulations. \citet{Tomasi2025} also used residual diffusion to downscale ERA5 to a high-resolution reanalysis fields of 2m temperature and 10m horizontal wind for Italy and found the diffusion approach outperformed baselines including a UNet and GAN. \citet{brazidec2026} also used residual diffusion along with a graph transformer encoder for global weather forecast downscaling from 100km to 30km. \citet{lopez-gomez2025} used a probabilistic diffusion model to downscale 45km resolution regional climate projections to 9km for an area in the US, for multiple variables, including precipitation. Recent work has shown that diffusion models consistently achieve better performance than traditional CNN and GAN approaches for downscaling precipitation, particularly in the capture of extremes \citep{preprint_fathom}. 

Unlike classical unconditioned diffusion models \citep{Ho2020}, climate downscaling requires learning a conditional distribution $p(y|x)$, where $y$ is the high-resolution target and $x$ represents coarse-resolution atmospheric and physiographic predictors. The choice of conditioning mechanism and how the model incorporates predictor information can significantly impact downscaling performance. These mechanisms are well established in the broader machine learning literature; for example, \citet{Rombach2022} demonstrated the effectiveness of cross-attention for incorporating information such as text prompts into generative models, and \citet{rozet2023} used conditioning via guidance applied at inference time. However, for climate downscaling, choice of conditioning mechanism is to date relatively unexplored.

Recent advances in deep learning have introduced AI models as emulators for weather forecasting. \citet{koldunov2024} demonstrated that AI NWP models trained on ERA5 data could be used to reproduce the correct features as that scale when provided with lower resolution inputs, thereby acting as a downscaling and bias correction tool. Pretrained foundation models that unify multiple downstream tasks, including downscaling, have been proposed. Models such as ClimaX \citep{nguyen2023climax} and Prithvi WxC \citep{schmude2024prithviwxcfoundationmodel} demonstrate that transformer-based architectures pretrained on global reanalysis or climate projections can be effectively fine-tuned for high-resolution downscaling. Similar capabilities have been shown by ORBIT-2 \citep{wang2025orbit2} and AtmoRep \citep{lessig2023atmorep}, which leverage specialized architectural adaptations and multimodal training, respectively.  Beyond deterministic approaches, generative foundation models like Climate in a Bottle (cBottle) \citep{climate_in_a_bottle} explicitly address probabilistic reconstruction and physical consistency at kilometer scales. These developments are supported by benchmarking frameworks such as ClimateLearn \citep{nguyen2023climatelearn}, which facilitate standardized evaluation. Together, these works signal a shift toward general-purpose foundation models for downscaling, while highlighting ongoing challenges in physical consistency and multivariate coupling.

Key challenges remain for weather foundation models, particularly in downscaling applications. Most models are still trained at coarse resolution, with high-resolution refinement handled externally and without standardized evaluation benchmarks. Physical inconsistencies, including violations of conservation laws and mismatches across variables, are frequently observed in super-resolved outputs. Extremes remain difficult to capture, and generalization across data sources (e.g., reanalysis to climate simulations) often requires substantial adaptation. High computational costs limit reproducibility, while limited interpretability, incomplete multi-modal integration, and restricted model availability continue to constrain broader operational use.

Despite recent progress in diffusion and foundation model approaches for climate downscaling, several important gaps remain. First, most diffusion-based downscaling studies rely on simple concatenation of conditioning variables, providing limited control over how large-scale atmospheric information influences fine-scale stochastic generation. Second, while weather foundation models have shown promise for deterministic downscaling, their potential as conditioning encoders for probabilistic generative models has not been well explored. Third, the computational and data-efficiency trade-offs associated with different conditioning strategies remain poorly quantified.

In this work, we address these limitations by evaluating alternative conditioning mechanisms for diffusion-based precipitation downscaling. Our contributions are threefold:

\begin{enumerate}
    \item We provide a controlled comparison of conditioning strategies, contrasting channel concatenation with cross-attention conditioning to assess their impacts on spatial structure, distributional realism, and extreme event representation.
    \item We introduce the use of a pretrained weather foundation model (Prithvi WxC) as a conditioned encoder within a probabilistic diffusion framework, demonstrating how pretrained representations can enhance generative performance.
    \item We quantitatively analyse how different conditioning approaches affect the amount of training data required to achieve comparable performance.
\end{enumerate}

To enable this, we evaluate channel concatenation-based conditioning against cross-attention conditioning using both a task-specific learnable convolutional encoder and a pretrained weather foundation model (Prithvi WxC), under identical training and inference settings. Performance is assessed using metrics spanning distributional accuracy, multiscale spatial structure, probabilistic calibration, extreme-event behavior, and data efficiency. 

\section{Dataset}
In this work, we focus on downscaling daily total precipitation in the Colorado River Basin, USA. In this context, diffusion models are trained using paired datasets consisting of (1) a high-resolution target variable (e.g. precipitation) from observations or reanalysis, and (2) a set of low-resolution conditioning climate variables (typically large scale meteorological variables). 

The target variable was obtained from ERA5-Land \citep{MunozSabater2019_ERA5Land} at 0.1°, extracted from the Copernicus Data Store (CDS)~\citep{C3S_ERA5_Land_2019}. The dynamic conditioning variables used for downscaling were taken from the 0.25° ERA5 reanalysis~\citep{Hersbach_et_al_2023}, also extracted from the CDS~\citep{Copernicus_C3S_ERA5_2023}. Consistent with established methods for statistical downscaling, precipitation was not included as a conditioning variable. Instead, we used large-scale atmospheric state variables associated with precipitation processes. Specifically, we used five vertical variables at three pressure levels and six surface variables, as summarized in Table \ref{tab:conditioning_variables}. 
All variables were extracted for the domain of 30°N to 46°N and 116°W to 100°W and aggregated (mean or total, as appropriate for the variable) to daily values. Predictor variables were coarse-grained to a 1° grid to match the typical spatial resolution used in climate projections. 

In addition to the dynamic variables we included five high-resolution static variables in the conditioning dataset. Three of these: \textit{cl}, \textit{lsm} and \textit{z} were taken directly from ERA5-land at 0.1° resolution. The remaining variables, \textit{sdor} and \textit{slor}, which are not available in the ERA5-Land dataset, were derived from a 0.01° digital elevation model based on Copernicus GLO-90 DEM~\citep{CopernicusDEM} and interpolated to the same ERA5-Land grid.

\begin{table}[H]
\caption{ERA5 atmospheric, surface, and static variables used to condition the diffusion precipitation downscaling models. Vertical atmospheric variables are provided at three pressure levels (500, 700, and 850 hPa). Static physiographic variables are provided at their native resolution, while surface and vertical variables are coarse-resolution.}

\centering
\small
\begin{tabular}{l l}
\toprule
\textbf{Variable} & \textbf{Description} \\
\midrule
\multicolumn{2}{l}{Vertical Atmospheric Variables at $\{500, 700, 850\}$ hPa} \\
\midrule
q & Specific Humidity \\
t & Temperature \\
u & Zonal Wind Speed (West--East) \\
v & Meridional Wind Speed (South--North) \\
w & Vertical Velocity \\
z & Geopotential \\
\midrule
\multicolumn{2}{l}{Surface Variables} \\
\midrule
cape & Convective Available Potential Energy \\
sp   & Surface Pressure \\
ssrd & Surface Solar Radiation Downward \\
t2m  & 2\,m Temperature \\
u10  & 10\,m Zonal Wind Speed \\
v10  & 10\,m Meridional Wind Speed \\
\midrule
\multicolumn{2}{l}{Static Physiographic Variables} \\
\midrule
cl   & Lake Cover \\
lsm  & Land Sea Mask \\
sdor & Standard Deviation of Subgrid-Scale Orography \\
slor & Slope of Subgrid-Scale Orography \\
z    & Surface Geopotential \\
 
\bottomrule
\end{tabular}
\label{tab:conditioning_variables}
\end{table}

\section{Methods}

This section describes the stochastic precipitation downscaling framework used in this study. We first introduce the conditional diffusion formulation and the base denoising architecture shared across all experiments. We then describe three alternative strategies for conditioning the diffusion model on coarse-resolution atmospheric and high-resolution static physiographic predictors: (1) channel concatenation-based conditioning (CC); (2) cross-attention conditioning using a task-specific learnable convolutional encoder (CA-CE); and (3) cross-attention conditioning with the frozen encoder of a pre-trained weather foundation model, Prithvi WxC (CA-PWC). These conditioning approaches represent the main experiments explored in this study. Finally, we describe the experimental setup used to evaluate the models.

\subsection{Conditioning Diffusion Models for Climate Downscaling}

Classical diffusion models define an unconditional generative process for a target high-resolution field $y$ by progressively adding noise to training data and learning to reverse this process~\citep{Ho2020}. However, in climate downscaling, we aim to learn a conditional distribution $p(y|x)$, where $x$ represents coarse-resolution predictors such as atmospheric variables and static physiological information. The conditioning process transforms the task from unconditional synthesis to the generation of high-resolution states that are physically-plausible weather samples~\citep{DiffESM}. 

We adopt the EDM formulation of diffusion models \citep{karras}, in which the forward process is defined by a continuous noise scale, $\sigma$ rather than a fixed discrete timestep schedule. Training data are perturbed with noise sampled across a wide range of magnitudes, and the model is trained to recover the clean signal at each noise level. The reverse process generates the high-resolution target variable: the model is initialized with noise along with the conditioning inputs and progressively denoises across a discretized sequence of noise levels to produce the target variable conditioned on the climate information.

We employ a UNet denoiser adapted for conditional generation, following the EDM preconditioning and loss-weighting scheme. This architecture integrates fine-scale local features with large-scale global context through its encoder–decoder structure and skip connections. Following \cite{karras}, the noise level, $\sigma$ is supplied to the network as a continuous conditioning input rather than a discrete timestep index, allowing the model to modulate its denoising behavior smoothly across the full range of noise magnitudes used during training (see Section \ref{sec:experiments}).

\subsection{Concatenation-based Conditioning}
\label{sec:concat}

\begin{figure*}

    \centering
    \includegraphics[width=1.05\linewidth]{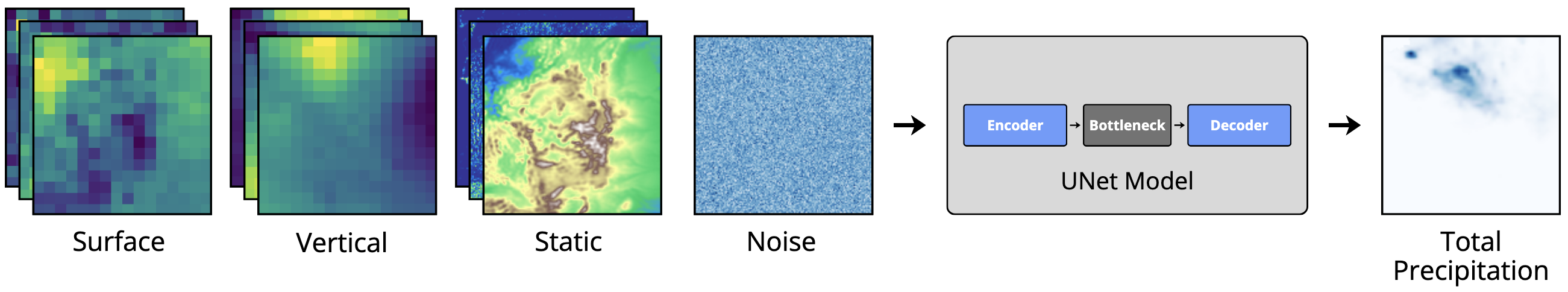}
    \caption{Schematic of concatenation-based conditioning (CC) for the diffusion precipitation downscaling model. Surface, vertical, and static physiographic predictor fields are concatenated with the noise input before being passed to the UNet denoising model, which outputs the downscaled total precipitation field. Surface and vertical coarse-resolution predictors must be interpolated to the high-resolution input grid.}
    \label{fig:concat_diagram}
\end{figure*}

Concatenation provides a straightforward and computationally efficient mechanism for incorporating predictor fields into diffusion models, as illustrated in Fig.~\ref{fig:concat_diagram}. In this approach, the conditioning variables are appended as additional channels to the noise input, a strategy widely used in conditional generative models and early diffusion frameworks~\citep{SR3, Ho2020} and widely used in climate downscaling~\citep{Addison}. This allows the model to directly access relevant atmospheric information and spatially detailed surface characteristics during the denoising process.

Although concatenation is often effective, it incorporates information implicitly by treating predictor fields as additional input channels, and the model must therefore learn how to extract and use this information from the raw inputs. As more variables are included, this approach can also become inefficient. Moreover, because the diffusion network operates on high-resolution inputs, this strategy requires upsampling the low-resolution predictors (Surface and Vertical variables) using interpolation before concatenation~\citep{srivastava2024precipitation}.

To incorporate coarse-resolution atmospheric and high-resolution physiographic predictors into the diffusion process, the model must be conditioned in a way that preserves large-scale context while maintaining fine-scale stochastic variability. Different conditioning mechanisms impose different inductive biases on how predictor information influences the denoising dynamics, with important implications for spatial structure, extremes, and uncertainty representations in downscaled precipitation.

\subsection{Cross-Attention Conditioning}

Cross-attention conditioning provides a more expressive mechanism for integrating coarse-resolution predictors into diffusion models than simple concatenation~\citep{cross_Attention_inductive_bias}. This approach allows the denoising network to dynamically query conditioning fields and selectively attend to the most relevant features at each denoising step, providing an effective modeling of the spatial and dynamical relationships critical for downscaling~\citep{10377753}. In contrast to concatenation-based conditioning, this approach removes the need for naive upsampling of coarse predictors. Although cross-attention has proven highly effective in computer vision applications such as text-to-image generation~\citep{survey_diffusion, Saharia}, its use in climate downscaling remains relatively underexplored and is not yet a standard conditioning strategy.

In this conditioning approach, coarse-resolution predictors must first be mapped into a latent embedding space~\citep{Rombach2022}. The cross-attention mechanism computes a context-aware representation that injects conditioning information directly into the latent feature maps of the denoiser model. Specifically, in cross-attention conditioning intermediate UNet feature maps act as \emph{queries}, while embeddings derived from coarse-resolution predictors (or other conditioning information) provide the \emph{keys} and \emph{values}. 

Formally, let $F \in \mathbb{R}^{N \times d}$ denote a UNet feature map flattened over spatial dimensions, and let $C \in \mathbb{R}^{M \times d_c}$ denote the conditioning embeddings. Linear projections are applied to obtain the \emph{queries}, \emph{keys}, and \emph{values}:
\begin{equation}
Q = F W_Q, \quad
K = C W_K, \quad
V = C W_V
\end{equation}
\noindent where $W_Q \in \mathbb{R}^{d \times d_k}$, $W_K \in \mathbb{R}^{d_c \times d_k}$, and $W_V \in \mathbb{R}^{d_c \times d_v}$ are learnable projection matrices.
The cross-attention operation is then defined as
\begin{equation}
\text{CrossAttention}(Q, K, V)
= \text{Softmax}\!\left(\frac{QK^\top}{\sqrt{d_k}}\right)V
\end{equation}
When predictors are provided as gridded fields or images, the embeddings can be obtained using an encoder model that produces a compact representation suitable for attention interactions. The resulting predictor embeddings are supplied to cross-attention blocks throughout the UNet architecture after each convolutional block at different spatial resolutions, allowing the denoising model to use the conditioning information to guide the generation of both large-scale contextual information and fine-scale features during the denoising process. 

\begin{figure*}
    \centering
    \includegraphics[width=0.8\linewidth]{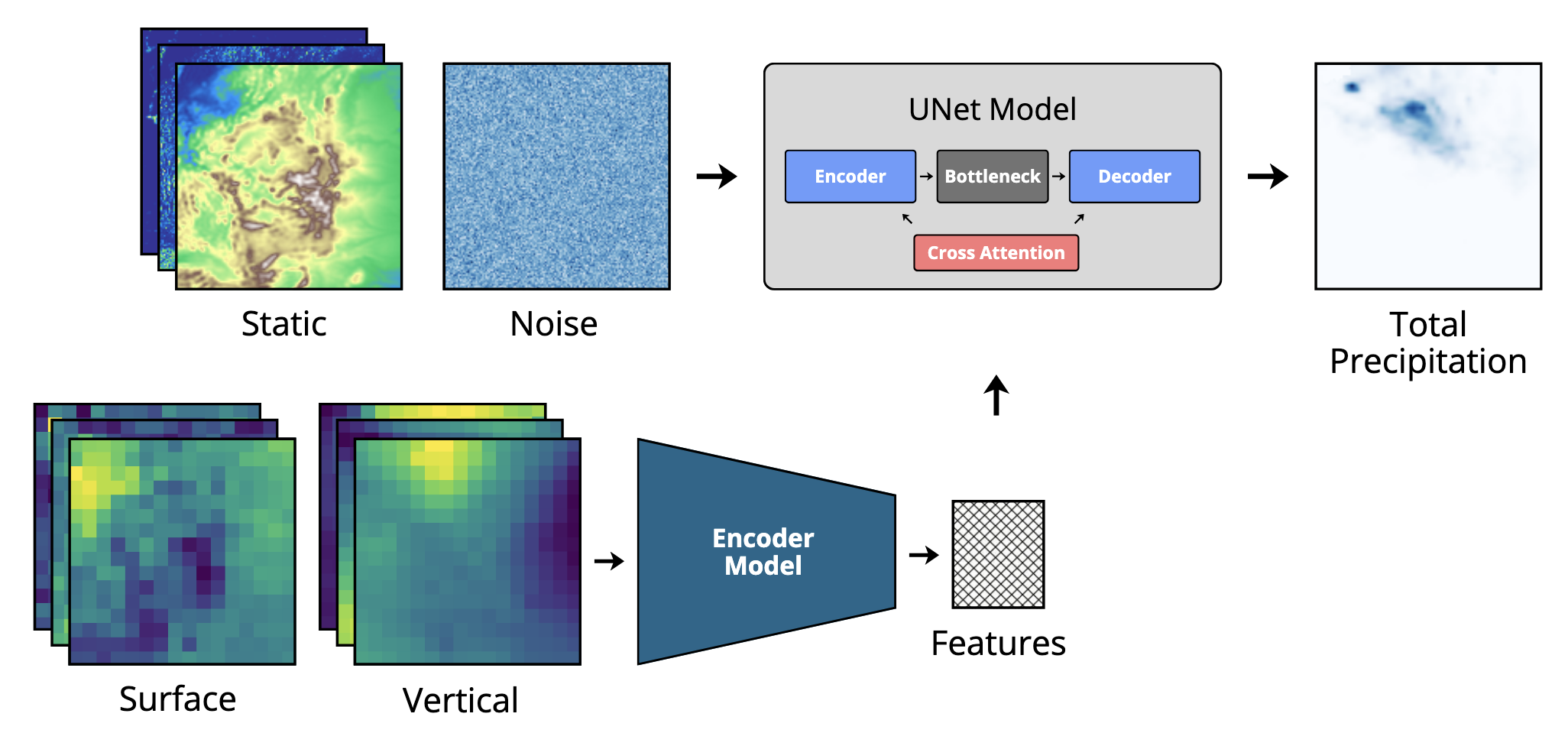}
    \caption{Schematic of cross-attention conditioning (CA) integrated into the UNet diffusion denoiser. Static physiographic variables are incorporated via channel concatenation at their native high resolution. Coarse-resolution surface and vertical predictors are encoded into a compact latent representation, which is supplied to cross-attention blocks at multiple spatial resolutions within the UNet encoder–decoder.}
    \label{fig:cross_diagram}
\end{figure*}

Because the encoder model operates on low-resolution inputs, we therefore retain static physiographic variables at their native high resolution and incorporate them via concatenation-based conditioning. In order to apply the same cross-attention mechanism to these static variables, it would require spatial coarsening, leading to a substantial loss of fine-scale information that is critical for downscaling. A high-level schematic of this conditioning strategy is shown in Figure~\ref{fig:cross_diagram}.

\subsection{Conditioning using Prithvi WxC Weather Foundation Model}

Recent advances in weather and climate AI led to the development of weather foundation models trained on massive Earth system datasets. Among these, the Prithvi WxC model~\citep{schmude2024prithviwxcfoundationmodel} provides a high-capacity spatiotemporal encoder specifically optimized for atmospheric fields. It was pretrained on Modern-Era Retrospective analysis for Research and Applications, Version 2 (MERRA-2) dataset~\citep{MERRA2}, a widely-used reanalysis dataset from NASA providing a broad range of global atmospheric and land surface spanning multiple decades.

The Prithvi WxC model follows a vision transformer architecture composed of hierarchical Swin Transformer blocks operating on 3D patches (latitude, longitude, and time). During pretraining, the model is exposed to decades of global data through a masked autoencoder objective, with the goal of learning rich multiscale representations of atmospheric dynamics. As a result, Prithvi WxC produces latent embeddings that capture both local- and global-scale features relevant for climate predictions. 

\begin{figure*}
    \centering
    \includegraphics[width=0.8\linewidth]{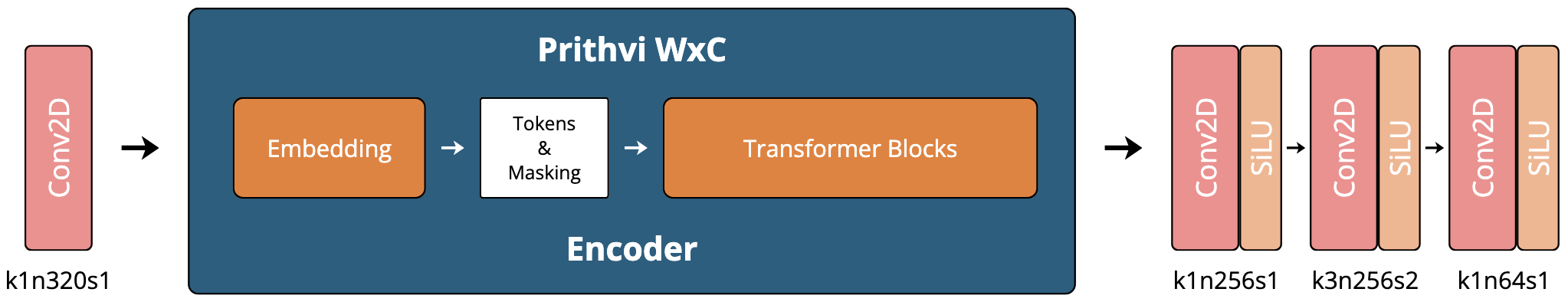}
    \caption{Modifications for the integration of the Prithvi WxC weather foundation model as a conditioning encoder within the diffusion framework. A lightweight convolutional preprocessing stage maps the coarse-resolution ERA5 predictor stack to the input format expected by the Prithvi WxC encoder. The frozen foundation model encoder produces high-dimensional spatiotemporal embeddings, which are subsequently projected to a compact feature map via convolutional post-processing layers and injected into the diffusion UNet via cross-attention.}
    \label{fig:prithvi_diagram}
\end{figure*}

To incorporate Prithvi WxC as a conditioning module within a diffusion downscaling model, we use only its encoder and add some extra convolutional layers, as illustrated in Figure~\ref{fig:prithvi_diagram}.

First, the raw predictors (e.g., coarse-resolution ERA5 variables) must be mapped to the input channel configuration expected by the Prithvi WxC encoder. We implement a lightweight convolutional preprocessing stage that converts the predictor stack into the format expected by the foundation model: a feature map with 320 channels. This ensures compatibility with the pretrained patch-embedding layer and avoids retraining the early stages of the transformer from scratch, which in our testing led to worse results. All Prithvi WxC parameters remain frozen during training.

Second, because the native Prithvi WxC output embeddings are substantially higher-dimensional than those typically used in UNet denoisers, leading to a high computational cost, a post-processing projection is applied. A convolutional head reduces the embedding dimensionality to produce a compact feature map suitable for injection into the diffusion denoiser. These reduced feature maps are then supplied to the cross-attention layers at multiple UNet spatial resolutions as described in the previous subsection. 

In summary, the frozen component is the pretrained Prithvi WxC encoder, which acts purely as a fixed feature extractor. The trainable components are: the convolutional preprocessing adapter, the convolutional post-processing projection head, and the diffusion UNet itself. This design conditions the diffusion model on latent representations derived from a weather foundation model, encoding physically meaningful features, while restricting gradient updates to the lightweight adapters and the denoiser. In practice, this leads to a form of semantically informed conditioning: the Prithvi WxC encoder supplies context about the large-scale patterns, while the diffusion model focuses on reconstructing high-resolution precipitation fields consistently with this context. The result is a conditioning mechanism designed to exploit foundation model representations in climate downscaling. 

To compare with the Prithvi WxC encoder (CA-PWC), we employ a learnable convolutional encoder (CA-CE) trained from scratch to map the coarse-resolution predictor fields into a compact latent representation. This encoder consists of three convolutional blocks, each comprising two 2D convolutional layers with SiLU activations. In each block, the first and third convolution preserves spatial resolution and is followed by feature-wise layer normalization to stabilize training, while the second convolution applies a stride of two to progressively downsample the feature maps, which increases the receptive field as each unit in deeper layers corresponds to a larger region of the original input. The channel dimensionality is kept constant across blocks. The final feature map is subsequently used as conditioning (\textit{keys} and \textit{values}) for the cross-attention layers of the diffusion UNet. This design provides a comparable and computationally efficient conditioning mechanism.

\subsection{Experiments}\label{sec:experiments}

As precipitation is a particularly challenging variable for statistical downscaling due to its high spatial heterogeneity and complex distributional tails, we employ a multi-faceted validation approach that evaluates the fields across several dimensions of quality. 

We evaluate the proposed stochastic precipitation downscaling framework following established practice in climate downscaling. We employ non-overlapping temporal splits for training, validation, and testing: the training set covers 1985--2009, the validation set covers 2010--2012, and the test set covers 2013--2015. This partition prevents leakage from future years during model selection and ensures comparability across all experiments. All input variables are normalized using scale factors computed from the training set.

To address the three research questions posed in the Introduction, model performance is assessed across four complementary dimensions. First, to evaluate whether cross-attention conditioning reproduces more realistic fine-scale spatial properties than concatenation, we quantify spectral fidelity via the Radially Averaged Power Spectral Density (RAPSD) and log-spectral distance (RALSD), and spatial skill via the Fractions Skill Score (FSS) at multiple intensity thresholds. Second, to assess whether foundation model embeddings improve distributional accuracy and extreme-event representation, we compute distributional fidelity metrics (distribution bias, histogram CRPS), annual maximum precipitation statistics (CRPS, MSE, MAE, and distribution bias on annual maxima), and event frequency ratios across intensity bins. Probabilistic accuracy (pixel-wise and spatially aggregated CRPS) and bias metrics (absolute bias, relative bias, and their time-averaged counterparts) provide a broader baseline of downscaling skill across all conditioning strategies; ensemble calibration is evaluated via the rank histogram calibration error. Third, to quantify the extent to which foundation model conditioning reduces dependency on large historical training datasets, we compare both cross-attention encoders trained on progressively shorter historical periods spanning 5 to 25 years. Formal definitions of all metrics are provided in \ref{app:metrics}.

All experiments employ a UNet denoising architecture with a symmetric encoder–decoder design consisting of four downsampling and four upsampling blocks. Skip connections between corresponding encoder and decoder blocks preserve information during the denoising process. The model is conditioned on diffusion timestep information using a Fourier time embedding, which modulates the denoising behavior across noise levels. The experiments using Prithvi WxC maintain the pretrained backbone of the foundation model frozen. 

As described in Section~\ref{sec:concat}, the concatenation approach requires upscaling the coarse ERA5 predictors to the high-resolution predictand grid, for this we used nearest-neighbour interpolation. The static variables are already available at high-resolution, so no additional upscaling is necessary. We also evaluate an unconditioned diffusion (UNC) model which serves as a baseline to isolate the architectural performance of the model from the gains provided by conditioning. Although it may capture broad statistical realism, it cannot accurately downscale specific weather events and is expected to fail on all conditional accuracy metrics.

During training, we apply log-uniform sampling of noise magnitudes following~\citet{climate_in_a_bottle}, with noise levels ranging from $\sigma_{\min}=0.02$ to $\sigma_{\max}=200.0$. This range is chosen to span the full dynamic range between the lowest-variance and highest-variance modes of the precipitation fields. The authors show that sampling sufficiently large noise levels during training is critical for diffusion models to generate realistic predictions. The diffusion model is trained for 100 epochs on a single NVIDIA A100 GPU using mixed-precision arithmetic with the bf16 floating-point format. To ensure fair comparisons between experiments, both training and inference are conducted using fixed random seeds, ensuring reproducibility across runs.

At inference time, stochastic high-resolution precipitation fields are generated using the noise discretization scheme introduced by~\citet{karras}. We employ $N=20$ sampling steps with $\rho=7$, which controls the spacing of noise levels $\{\sigma_{\max}, \ldots, \sigma_{\min}\}$ in the discretization schedule, followed by a final denoising step with $\sigma=0$. This sampling strategy enables a gradual transition from coarse stochastic structure to fine-scale spatial refinement. For each day in the test period, we generate five stochastic realizations for evaluation.

Overall, this experimental design, comprising standardized climate data, strict temporal splits, log-uniform noise sampling during training, a 20-step sampling schedule for inference, and 5-member ensemble predictions, provides a robust and reproducible setup for assessing stochastic diffusion-based precipitation downscaling.

\section{Results and Discussion}

We evaluate the proposed diffusion models for precipitation downscaling by comparing the four conditioning strategies under identical training, inference, and evaluation settings. All quantitative results are reported for daily total precipitation (mm/day) over the held-out test period (2013--2015), with five stochastic ensemble members generated per test day, ensuring that performance differences arise solely from the conditioning mechanism rather than data leakage or architectural variation.

\subsection{Precipitation Evaluation}
Here we present the results of the evaluation of pixel-level accuracy of the generated spatial rainfall time series, alongside distributional accuracy/realism, including the ability to reproduce extreme events. For daily total precipitation (Table~\ref{tab:statistical_metrics}), pointwise and spatially aggregated error metrics show that the CC model achieves the lowest raw CRPS (0.511) and MSE (4.282), indicating superior pixel-level accuracy. The CA-CE and CA-PWC models perform similarly to each other (raw CRPS of 0.581 and 0.557, respectively), while the UNC baseline exhibits substantially larger errors across all metrics as expected. However, when considering bias measures, CA-PWC achieves the lowest time-averaged absolute bias (0.095 mm/day) and absolute relative bias (0.097), outperforming both CC (0.185, 0.169) and CA-CE (0.107, 0.112). 

Precipitation distribution and spatial spectral metrics (Table~\ref{tab:spatial_metrics}) indicate superior distributional performance of the cross-attention models. CA-PWC achieves the lowest distribution bias (0.065), nearly five times lower than CC (0.385) and half of the CA-CE (0.138). Despite slightly higher calibration error relative to the UNC model, both cross-attention models produce more accurate spatial power spectra, the CA-CE and CA-PWC achieve RAPSD CRPS values of 2.632 and 2.598, compared with 2.763 for CC and 7.011 for UNC. RALSD scores follow the same ordering, with CA-CE (5.390) and CA-PWC (5.539) outperforming CC (5.942), suggesting that cross-attention conditioning better preserves fine-scale spatial structure, producing realistic spatial rainfall distributions.

\begin{table}[H]
\centering
\caption{Comparison of conditioning strategies for probabilistic accuracy and bias metrics for daily total precipitation (mm/day) over the test period (2013–2015). Bold values indicate best performance per metric. "Raw" values correspond to mean of daily per-pixel precipitation metrics. "x16" are calculated on 16x16 degree grid mean values. and "Time avg" on mean values across all time steps in the test period. }\label{tab:statistical_metrics}
\setlength{\tabcolsep}{5pt}
\begin{tabular}{l|ccccccccc}
\toprule
& \multicolumn{2}{c}{\textbf{CRPS}} & \multicolumn{2}{c}{\textbf{MSE}} & \multicolumn{2}{c}{\textbf{Abs. Bias}} & \multicolumn{2}{c}{\textbf{Abs. Rel. Bias}} \\
\cmidrule(lr){2-3} \cmidrule(lr){4-5} \cmidrule(lr){6-7} \cmidrule(lr){8-9}
 & Raw & (x16) & Raw & (x16) & Raw & Time avg. & Raw & Time avg. \\
\midrule
UNC       & \underline{1.092} & \underline{1.034} & \underline{15.993} & \underline{11.893} & \underline{1.617} & \underline{0.402} & \underline{317.22} & \underline{0.380} \\
CC       & \textbf{0.511} & \textbf{0.389} & \textbf{4.282}  & \textbf{1.918}  & \textbf{0.760} & 0.185 & 45.656 & 0.169 \\
CA-CE       & 0.581 & 0.444 & 5.497  & 2.825  & 0.876 & 0.107 & 49.494 & 0.112 \\
CA-PWC & 0.557 & 0.415 & 5.934  & 2.548  & 0.827 & \textbf{0.095} & \textbf{35.710} & \textbf{0.097} \\
\bottomrule
\end{tabular}
\end{table}

\begin{table}[H]
\centering
\caption{Comparison of conditioning strategies for precipitation distribution and spatial spectral metrics on the test period.}\label{tab:spatial_metrics}
\begin{tabular}{l|ccccc}
\toprule
 & \multicolumn{3}{c}{\textbf{Precipitation Distribution}} & \multicolumn{2}{c}{\textbf{Spatial Statistics}} \\
\cmidrule(lr){2-4} \cmidrule(lr){5-6}
 & Distribution & Calibration & Histogram & RALSD & RAPSD \\
 & Bias & Error & CRPS & & CRPS \\
\midrule
UNC       & 0.358 & \textbf{0.028} & \underline{0.150} & \underline{13.107} & \underline{7.011} \\
CC       & \underline{0.385} & 0.065 & \textbf{0.102} & 5.942 & 2.763 \\
CA-CE       & 0.138 & \underline{0.079} & \textbf{0.102} & \textbf{5.390} & 2.632 \\
CA-PWC & \textbf{0.065} & 0.050 & \textbf{0.102} & 5.539 & \textbf{2.598} \\
\bottomrule
\end{tabular}

\end{table}

\begin{figure}[H]
    \centering
    \includegraphics[width=1.0\linewidth]{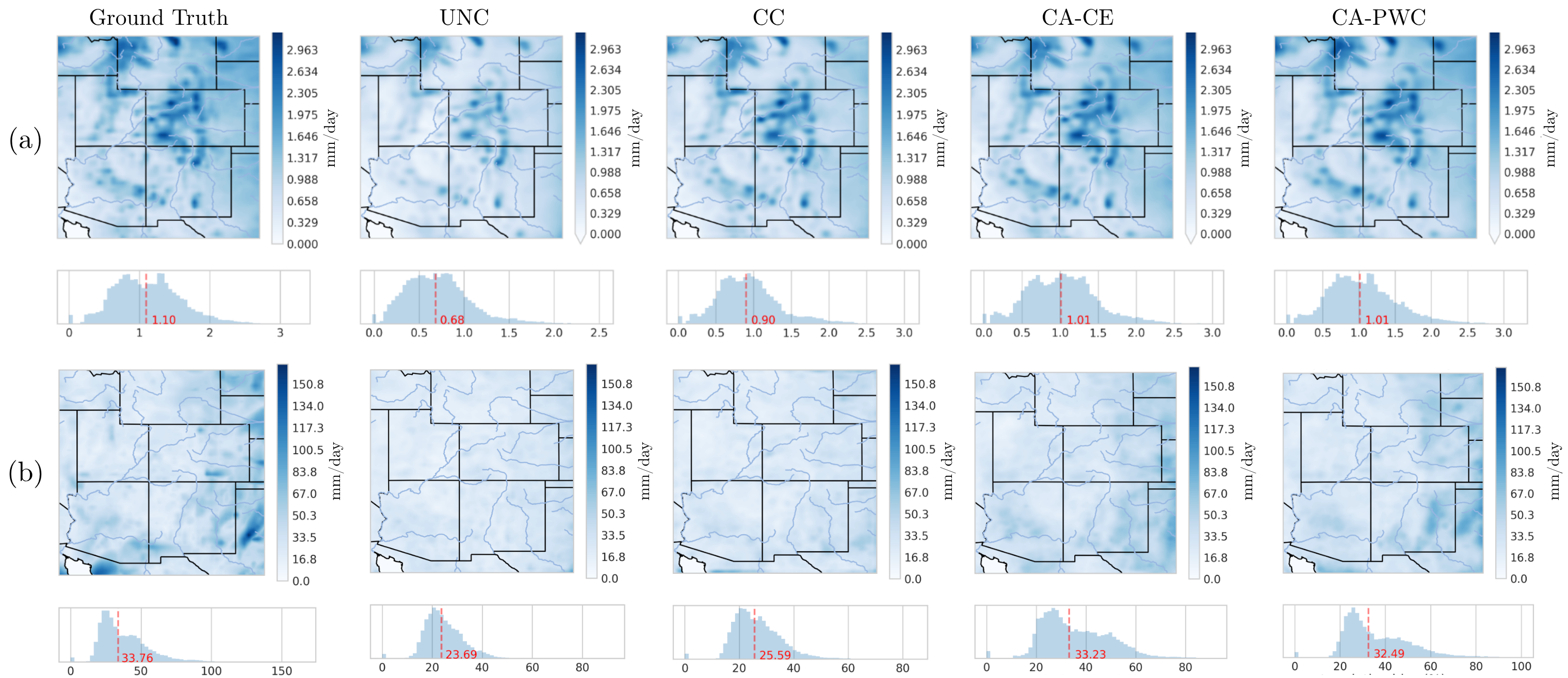}
    \caption{Mean and maximum daily total precipitation (mm/day) over the test period (2013–2015) for the ERA5-Land reference dataset and each conditioning strategy. Top row (a): time-mean daily precipitation fields and corresponding marginal histograms. Bottom row (b): time-maximum daily precipitation fields and corresponding marginal histograms. Red dashed lines on histograms indicate the reference mean value.}
    \label{fig:merged_daily_precipitation}
\end{figure}

\begin{figure}[H]
    \centering
    \includegraphics[width=1.0\linewidth]{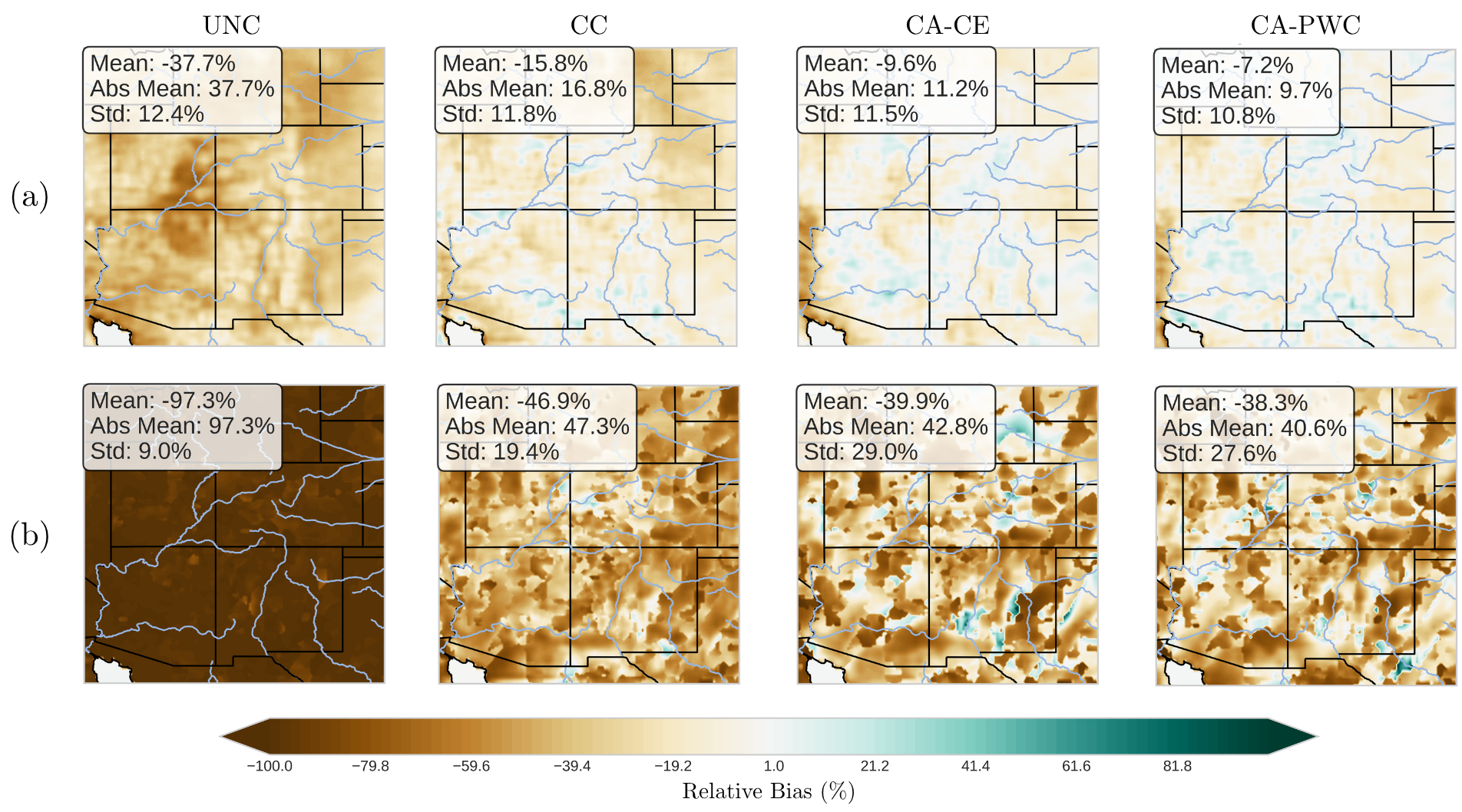}
    \caption{Spatial maps of mean relative bias (\%) in daily total precipitation (mm/day) for the test period (2013–2015). Top row: pixel-wise mean relative bias across all test days. Bottom row: mean relative bias computed after spatial maximum aggregation.}
    \label{fig:merged_spatial_mean_bias}
\end{figure}

This contrast between pixel-level accuracy and distributional fidelity reflects the difference in how conditioning strategies represent precipitation. Spatial maps of mean and maximum daily precipitation (Figure~\ref{fig:merged_daily_precipitation}) illustrate these differences visually. All three conditioned models reproduce the broad spatial patterns of mean precipitation, but the CC model produces noticeably smoother fields. The marginal histograms confirm that CC underestimates the frequency of high-intensity events in the test dataset, whereas CA-CE and CA-PWC more faithfully reproduce the reference right tail. This effect is amplified in the maximum precipitation fields, where the cross-attention models capture peak values closer to the reference. Spatial bias maps (Figure~\ref{fig:merged_spatial_mean_bias}) further show that CA-PWC exhibits the smallest mean relative bias across most of the domain for both daily and maximum precipitation. A notable shared feature of the CA-CE and CA-PWC models is a dry bias extending northward from the south-western corner of the domain. For maximum precipitation, the CA-PWC model tends to exhibit a dry bias to varying degrees across the domain, while the CA-CE also has a wet bias in some locations. 

The advantage of cross-attention conditioning is most pronounced for annual maximum precipitation (Table~\ref{tab:precipitation_metrics}). CA-CE achieves the lowest MSE (148.712) and MAE (8.271), while CA-PWC achieves both the lowest CRPS (5.730) and the lowest distribution bias (0.062) for annual maxima. In contrast, the CC model performs worse on all four metrics, with a distribution bias of 0.388 comparable to the UNC baseline (0.359). These results are consistent with the hypothesis that cross-attention conditioning better captures precipitation extremes, although we note that the evaluation is based on a relatively limited sample of three years and so this result should be treated with caution.

\begin{table} 
\centering 
\caption{Comparison of conditioning strategies for annual maximum daily precipitation (mm/day) over the test period (2013–2015). Bold values indicate best performance per metric.} 
\begin{tabular}{l|cccc} 
\toprule 
& \multicolumn{4}{c}{\textbf{Annual Maximum Precipitation}} \\
\cmidrule(lr){2-5}
& \textbf{CRPS} & \textbf{MSE} & \textbf{MAE} & \textbf{Distribution Bias} \\ 
\midrule 
UNC       & \underline{9.234} & \underline{288.551} & \underline{11.869} & 0.359 \\ 
CC       & 7.522 & 194.825 & 9.486 & \underline{0.388} \\ 
CA-CE       & 5.750 & \textbf{148.712} & \textbf{8.271} & 0.134 \\ 
CA-PWC & \textbf{5.730} & 175.959 & 8.854 & \textbf{0.062} \\ 
\bottomrule 
\end{tabular} 
\label{tab:precipitation_metrics} 
\end{table}

\begin{figure}[H]
    \centering
    \includegraphics[width=0.5\linewidth]{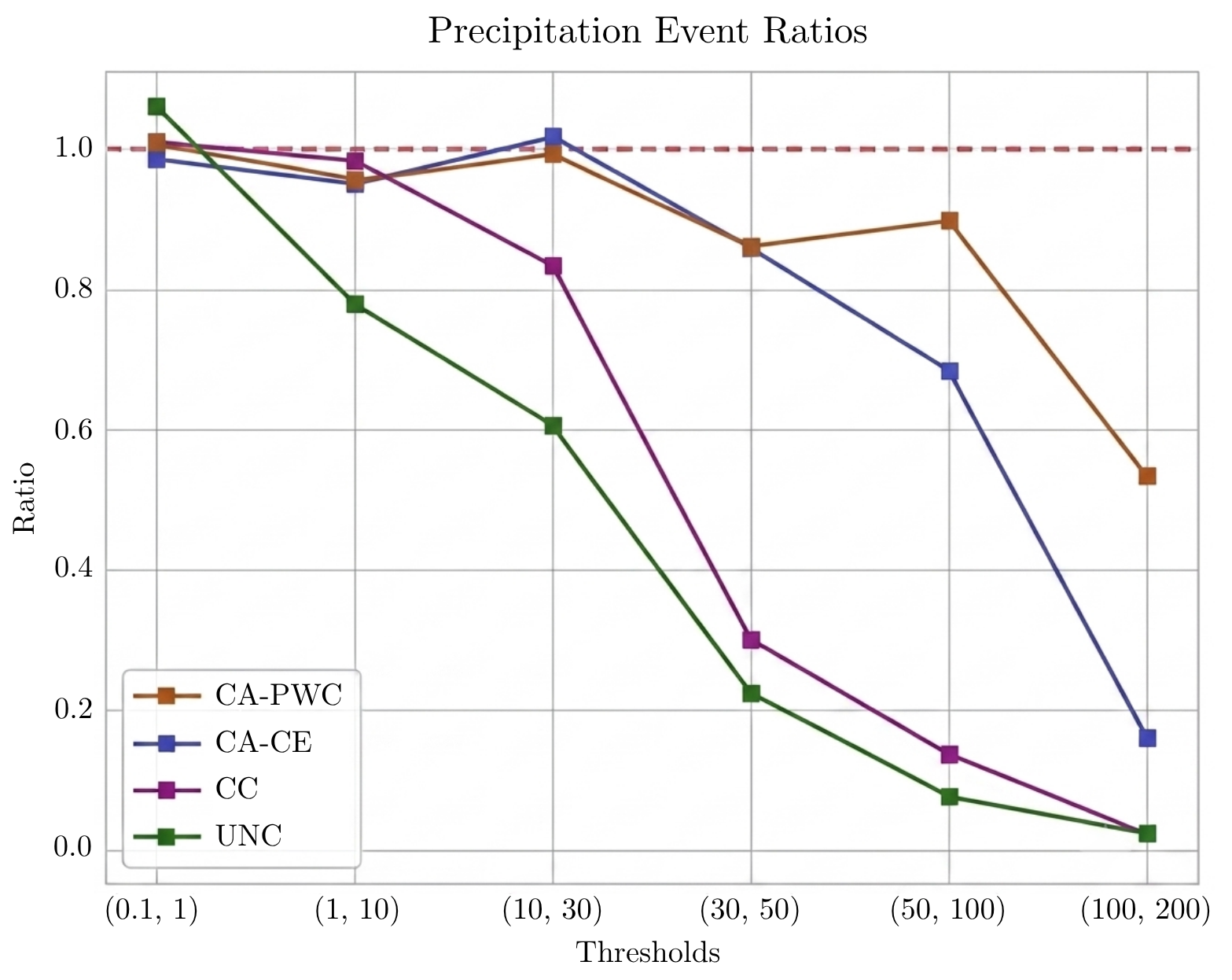}
    \caption{Event frequency ratios across precipitation intensity bins (mm/day) for each conditioning strategy over the test period (2013–2015). The ratio is computed as the number of predicted events divided by the number of reference events within each intensity bin, with a value of 1.0 (dashed red line) indicating a perfect match. Observation pixel-event counts: 0-0.1 mm/day (n=14.38M); 0.1-1 mm/day (n=4.68M); 1-10 mm/day (n=4.40M); 10-30 mm/day (n=569.6k); 30-50 mm/day (n=37.9k); 50-100 mm/day (n=6.7k); 100-200 mm/day (n=263). Lower counts in extreme bins indicate higher uncertainty in those ratios.}
    \label{fig:event_ratios}
\end{figure}

Precipitation event frequency ratios (Figure~\ref{fig:event_ratios}) provide a direct diagnostic of extreme-event fidelity. For events in the 100--200 mm/day range, CA-PWC retains over 50\% of reference event counts, CA-CE retains approximately 17\%, and the CC model produces virtually no events at these intensities (ratio $\approx$ 0). All models perform comparably for moderate intensities (below 20 mm/day), where the choice of conditioning strategy has little effect. 

\begin{figure}[H]
    \centering
    \includegraphics[width=0.5\linewidth]{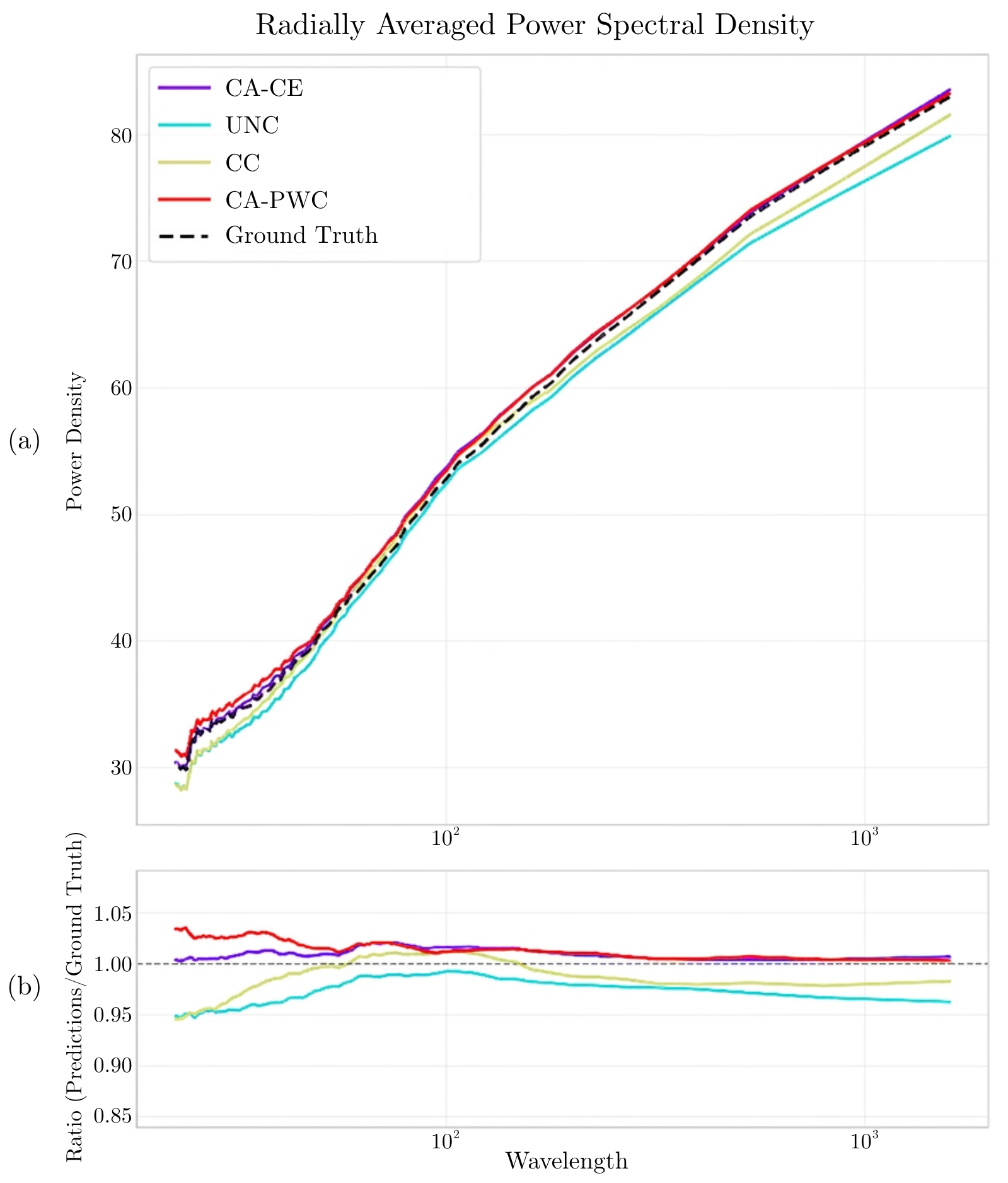}
    \caption{Radially averaged power spectral density (RAPSD) of daily total precipitation over the test period (2013–2015) for the ERA5-Land reference dataset and each conditioning strategy (top panel). The bottom panel shows the ratio of predicted to reference power spectra as a function of spatial wavelength (km). Values close to 1.0 indicate accurate reproduction of spatial variability at a given scale.}
    \label{fig:power_spectra}
\end{figure}

Radially averaged power spectral density curves (Figure~\ref{fig:power_spectra}) corroborate these findings, where both cross-attention models more accurately reproduce the observed spatial power spectrum at scales below approximately 100~km, where the CC model shows a decay in spectral power consistent with its tendency to produce over-smoothed outputs.

\begin{figure}[H]
    \centering
    \includegraphics[width=1.0\linewidth]{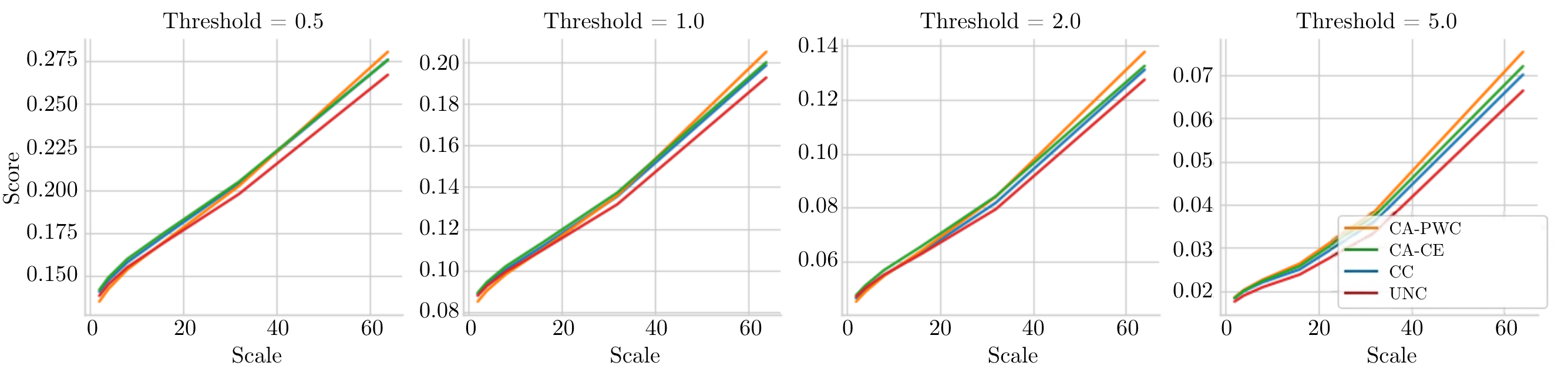}
    \caption{Fractions Skill Score (FSS) for daily total precipitation at multiple intensity thresholds (mm/day) as a function of spatial scale (km) for each conditioning strategy over the test period (2013–2015). Higher FSS values indicate better spatial skill at a given scale and threshold.}
    \label{fig:fss_score}
\end{figure}

Fractions Skill Score results (Figure~\ref{fig:fss_score}) indicate moderately better spatial skill for CA-PWC and CA-CE at higher intensity thresholds and finer spatial scales. For larger thresholds, CA-PWC generally achieves higher FSS than other models, while differences among models are small for low-intensity events. 

In summary, these results reveal a clear trade-off between pixel-level accuracy, where concatenation held strong performance, and distributional and extreme-event fidelity, where cross-attention conditioning with either a convolutional or foundation-model encoder offers substantial advantages across the distributional, spectral, and extreme event metrics considered here.

\subsection{Data Efficiency and Computational Cost}

\begin{figure}[H]
    \centering
    \includegraphics[width=1.0\linewidth]{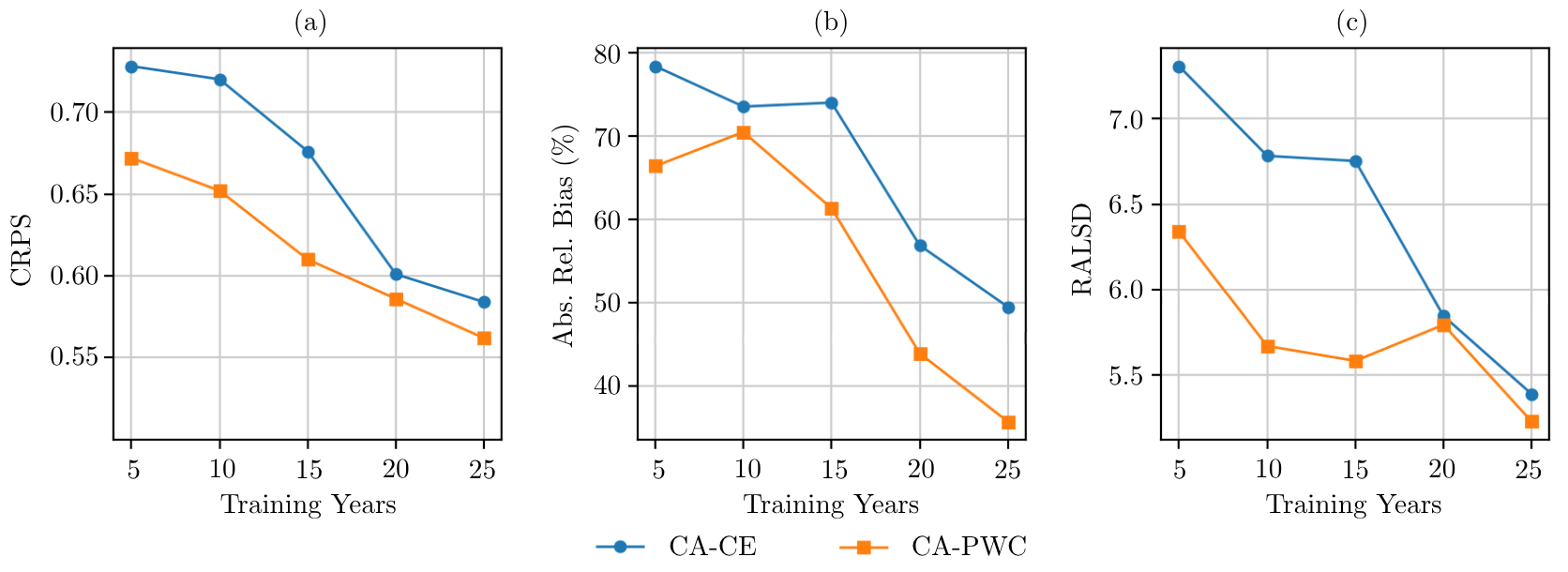}
    \caption{Data efficiency of cross-attention conditioning strategies as a function of training data volume (5--25 years) for CA-CE and CA-PWC. (a): CRPS, (b): Absolute Relative Bias, (c): RALSD.}
    \label{fig:data_efficiency}
\end{figure}

To assess how well each cross-attention encoder exploits limited training data, we compare CA-CE and CA-PWC trained on five progressively shorter historical periods (5, 10, 15, 20, and 25 years), evaluating on the fixed 2013--2015 test set. Figure~\ref{fig:data_efficiency} shows CRPS, time-averaged absolute relative bias, and RALSD as a function of training volume.

Across all three metrics CA-PWC consistently outperforms CA-CE for all experiments conducted in this study. For CRPS, both models improve with more data but remain above the Concatenation baseline (0.511) at all volumes, with CA-PWC degrading less steeply as data is reduced (16\% vs. 20\%). For absolute relative bias, both cross-attention models stay below the CC baseline (0.185) on high volumes of data. The clearest data-efficiency advantage emerges in RALSD, where Prithvi WxC falls below the Concatenation threshold (5.94) already at 10 years and reaches 5.23 at 25 years, while CA-CE only crosses that threshold at the 20-year period, suggesting that the pretrained spatial representations in Prithvi WxC, when combined with cross-attention conditioning, are particularly effective for learning precipitation structure from fewer training samples.

\begin{table}[H]
\centering
\caption{Model complexity across conditioning strategies. Trainable parameters refers to the number of parameters updated during training, while total parameters includes the frozen Prithvi WxC backbone. GPU memory usage includes loading both model and data. Inference time is reported per day per ensemble for the 1095 day $\times$ 5 ensemble test set on a single NVIDIA H100 80GB HBM3.}\begin{tabular}{l|ccccc}
\toprule
& \textbf{Trainable} & \textbf{Total} & \textbf{GPU Memory} & \textbf{Inference} & \textbf{Training Time} \\
& \textbf{Params}    & \textbf{Params} & \textbf{Usage} & \textbf{Time}      & \textbf{per Epoch} \\
\midrule
UNC  & 48.9 M & 48.9 M & 15.9 GB & 0.210 s & 138 s \\
CC  & 48.9 M & 48.9 M & 16.0 GB & 0.211 s & 141 s \\
CA-CE  & 52.9 M & 52.9 M & 27.8 GB & 0.605 s & 324 s \\
CA-PWC    & 85.0 M & 301 M  & 29.4 GB & 0.684 s & 342 s \\
\bottomrule
\end{tabular}
\label{tab:complexity}
\end{table}

Table~\ref{tab:complexity} summarises model complexity across all four conditioning strategies. The UNC and CC models share the same parameter count (48.9 M) since concatenation adds no learnable components beyond the base UNet denoiser. The CA-CE model adds 4 M parameters for the three-block convolutional encoder (52.9 M total). CA-PWC has 301 M total parameters, but only 85 M are trainable as the 216 M Swin-Transformer backbone is kept frozen, representing a $1.6\times$ increase in trainable parameters relative to CA-CE. Training cost is therefore still relatively modest despite the large backbone, though inference cost is higher as the full frozen backbone must be evaluated on every sample.

\section{Conclusions}

This paper evaluated conditioning strategies for diffusion-based statistical downscaling of precipitation over the Colorado River Basin, comparing an unconditioned baseline, a concatenation approach, and two cross-attention approaches using a learned convolutional encoder and the Prithvi WxC foundation model. All experiments were conducted under identical training and inference conditions, isolating the effect of the conditioning mechanism. The results reveal a consistent trade-off: concatenation conditioning achieved the lowest pixel-level CRPS and MSE, but cross-attention conditioning with either encoder produced more accurate precipitation distributions and improved representation of extreme events.

For daily precipitation, the CC model achieved the best raw probabilistic accuracy, outperforming the cross-attention models on CRPS and MSE at every grid point. However, this advantage is associated with a tendency of concatenation to produce oversmoothed fields and suppress high-intensity events. The cross-attention models achieved distribution biases up to five times lower than Concatenation, reproduced the spatial power spectrum more faithfully at high scales, and matched observed annual maximum precipitation distributions far more accurately, although annual maximum statistics were derived from only three years of observations and are therefore subject to substantial sampling uncertainty. These differences were most pronounced for extreme events: for rainfall intensities above 100 mm/day, CA-PWC retained over 50\% of observed event counts while the CC model produced virtually none, although again these results are uncertain due to the limited number of observed tail events.

CA-PWC and the CA-CE performed broadly similarly across most daily metrics. CA-PWC achieved the lowest CRPS and distribution bias for annual maxima, the lowest time-averaged absolute relative bias, and marginally better spatial spectral scores, whereas CA-CE achieved lower MSE and MAE on annual maximum precipitation and slightly better spatial realism metrics.
CA-PWC and CA-CE therefore provide two complementary alternatives to CC when distributional realism is important. CA-CE offers a lightweight, operationally practical approach, while CE-PWC can additionally address limited training data regimes. Despite having 301 M total parameters and a larger memory footprint, only 85 M parameters were trained in the CE-PWC model, meaning it was able to offer these data efficiency advantages with only a relatively small increase in training and inference times.

For this regional precipitation downscaling task, the results presented indicate that cross-attention conditioning improves distributional realism, spectral fidelity, and extreme-event representation relative to simple channel concatenation. For applications in climate impact assessment, where the frequency and intensity of extreme precipitation events are of primary concern, this may be particularly advantageous. 

This study is subject to several limitations. We considered only a single geographic domain, the Colorado River Basin, and a single variable, daily precipitation, therefore results may not generalise to other regions, climate regimes, or variables. In addition, all experiments rely on ERA5 reanalysis data for conditioning, which may not capture the biases and non-stationarities present in GCM climate projections. Similarly, we used a target of ERA5-Land, which does not directly integrate precipitation observations. Our evaluation period of three years means that metrics focused on the most extreme events, particularly annual maxima, are derived from a relatively small effective sample and are therefore associated with a greater uncertainty than those metrics derived from daily rainfall. While the training plus validation period of 28 years was sufficient for the CA models to learn representations of extreme events, it is possible that a concatenation-based model could narrow the gap in extreme-event performance if trained on a larger dataset. This would need further work to evaluate. The transferability of these findings to future climate scenarios remains uncertain, and warrants further evaluation across multiple basins, variables, and data sources. Future work should also consider whether jointly fine-tuning the foundation model backbone on regional data can further narrow the gap in pixel-level accuracy without sacrificing distributional fidelity.

\section*{Acknowledgments}
This work was funded by the Hartree Centre for Digital Innovation, a collaboration between IBM and STFC. Generated using or contains modified Copernicus Climate Change Service information, 2019. Neither the European Commission nor ECMWF is responsible for any use that may be made of the Copernicus information or data it contains.

\appendix 

\section{Appendix}
\subsection{Evaluation Metrics}
\label{app:metrics}

\textbf{Probabilistic Accuracy:} We use the Continuous Ranked Probability Score (CRPS) to quantify the accuracy of the 5-member ensemble predictions. The CRPS measures the distance between the cumulative distribution function (CDF) of the predicted ensemble $F(x)$ and the reference data $y$:
\begin{equation}
CRPS(F, y) = \int_{-\infty}^{\infty} (F(z) - \mathbb{I}(z \ge y))^2 dz
\end{equation}
where $\mathbb{I}$ is the indicator function. We report both the pixel-wise CRPS averaged over the domain and the CRPS of  spatially aggregated fields (x16) to assess performance at different scales.

In addition, we compute a histogram CRPS to evaluate distributional consistency. Reference data and ensemble members are transformed to log-space and converted into normalized histograms over fixed bins, which define discrete distributions. The CRPS is then computed between the observed histogram and the ensemble of predicted histograms. It provides a measure of how well the ensemble reproduces the overall value distribution independent of spatial structure.

\textbf{Distribution Bias:} To assess how well the model reproduces the climatological distribution of precipitation, we compute the Distribution Bias, defined as the integrated absolute difference between the empirical CDF of the model ($F_{mod}$) and the reference ($F_{ref}$) across all grid points and time steps:
\begin{equation}
\text{Distribution Bias} = \int |F_{mod}(x) - F_{ref}(x)| dx
\end{equation}
This metric captures systematic errors in the frequency of precipitation intensities, independent of temporal synchronization.

\textbf{Bias Metrics:} To assess systematic errors in the predicted precipitation fields, we compute several bias metrics operating at different levels of temporal aggregation. The absolute bias is defined as the mean absolute difference between predictions and reference across all ensemble members, time steps, and grid points:
\begin{equation}
\text{Absolute Bias} = \frac{1}{N} \sum_{t,i,j} |\hat{y}_{t,i,j} - y_{t,i,j}|
\end{equation}

The relative bias normalizes the absolute pointwise error by the observed value at each location, providing a scale-independent measure of error that is especially informative in regions of low mean precipitation where absolute errors are small but proportionally large. It is computed over all non-zero observed values  (totaling $N^*)$ to avoid division by zero:
\begin{equation}
\text{Absolute Relative Bias} = \frac{1}{N^*} \sum_{\substack{t,i,j \\ y_{t,i,j} \neq 0}} \frac{|\hat{y}_{t,i,j} - y_{t,i,j}|}{y_{t,i,j}}
\end{equation}

The time-averaged bias collapses the temporal dimension by first computing the mean predicted and observed fields across all time steps, and then evaluating their difference at each spatial location. This metric therefore quantifies bias in the climatological mean field, rather than errors at individual time steps, and highlights spatial discrepancies between the predicted and observed long-term averages.

\textbf{Calibration Error:} We evaluate the reliability of the ensemble spread using the Calibration Error, which measures the consistency between predicted probabilities and observed frequencies. For an ensemble of size $M$, if the reference falls into the $k^{th}$ bin of the predictive distribution with frequency $p_k$, the error is:
\begin{equation}
\text{Calibration Error} = \sum_{k=1}^{M+1} (p_k - \frac{1}{M+1})^2
\end{equation} Ideally, the rank histogram should be uniform, corresponding to a calibration error of zero.

\textbf{Spectral Fidelity:} Spatial realism is evaluated using the Radially Averaged Power Spectral 
Density (RAPSD). The RAPSD $S(k)$ at wavenumber $k$ is computed by azimuthally averaging the 2D power spectrum $|\mathcal{F}(I)(u,v)|^2$:
\begin{equation}
S(k) = \frac{1}{N_k} \sum_{k \le \sqrt{u^2+v^2} < k+1} |\mathcal{F}(I)(u,v)|^2
\end{equation}
where $\mathcal{F}$ denotes the Fourier transform and $N_k$ is the number modes in the annulus. To compare the texture of generated fields against the ERA5-Land reference, we compute the log-spectral distance (RALSD) between the model and reference spectra.

\textbf{Fractions Skill Score:} Spatial accuracy is evaluated using the Fractions Skill Score (FSS), which evaluates agreement between predicted ensembles and observed precipitation within a local neighborhood with a exceedance threshold $\tau$. For each member $m$, a binary exceedance field is constructed as $\mathbb{I}(\hat{y}_{t,i,j}^{(m)} \ge \tau)$, and the ensemble mean fractional coverage $\bar{f}_s(t,i,j)$ is obtained by averaging these fields over a window of size $s$:
\begin{equation} 
\text{FSS}(s, \tau) = 1 - \frac{\displaystyle\sum_{t,i,j} \left(\bar{f}_s(t,i,j) - f_s^{\text{obs}}(t,i,j)\right)^2}{\displaystyle\sum_{t,i,j} \bar{f}_s(t,i,j)^2 + \displaystyle\sum_{t,i,j} \left(f_s^{\text{obs}}(t,i,j)\right)^2}
\end{equation}
where $f_s^{\text{obs}}(t,i,j)$ is the corresponding fractional coverage of the binary reference field $\mathbb{I}(y_{t,i,j} \ge \tau)$ within the same window. The score ranges from 0 (no skill) to 1 (perfect agreement). We report the mean FSS averaged across all samples.

\bibliographystyle{plainnat}
\bibliography{refs}

\end{document}